\documentclass[10pt,twocolumn,letterpaper]{article}
\pdfoutput=1
\usepackage{wacv}
\usepackage{times}
\usepackage{epsfig}
\usepackage{graphicx}
\usepackage{amsmath}
\usepackage{amssymb}
\usepackage{subfigure} 
\usepackage[ruled,linesnumbered]{algorithm2e}
\usepackage{makecell}
\usepackage[titletoc]{appendix}

\wacvfinalcopy % *** Uncomment this line for the final submission

\def\wacvPaperID{302} % *** Enter the wacv Paper ID here

\ifwacvfinal\fi
\begin{document}

%%%%%%%%% TITLE
\title{CANZSL: Cycle-Consistent Adversarial Networks for Zero-Shot Learning from Natural Language}

% Authors at the same institution
%\author{First Author \hspace{2cm} Second Author \\
%Institution1\\
%{\tt\small firstauthor@i1.org}
%}
% Authors at different institutions
\author{Zhi Chen \\
School of ITEE, \\ University of Queensland\\
{\tt\small uqzhichen@gmail.com}
\and
Jingjing Li \\
University of Electronic Science \\ and Technology of China\\
{\tt\small lijin117@yeah.net}
\and 
Yadan Luo \\
School of ITEE, \\
 University of Queensland\\
{\tt\small lyadanluol@gmail.com}
\and
\and
Zi Huang \\
School of ITEE, \\
 University of Queensland\\
{\tt\small huang@itee.uq.edu.au}
\and
Yangyang \\
University of Electronic \\ Science and Technology of China \\
{\tt\small dlyyang@gmail.com}}

\maketitle
\ifwacvfinal\thispagestyle{empty}\fi

%%%%%%%%% ABSTRACT
\begin{abstract}
    Existing methods using generative adversarial approaches for Zero-Shot Learning (ZSL) aim to generate realistic visual features from class semantics by a single generative network, which is highly under-constrained. As a result, the previous methods cannot guarantee that the generated visual features can truthfully reflect the corresponding semantics. To address this issue, we propose a novel method named Cycle-consistent Adversarial Networks for Zero-Shot Learning (CANZSL). It encourages a visual feature generator to synthesize realistic visual features from semantics, and then inversely translate back synthesized the visual feature to corresponding semantic space by a semantic feature generator. Furthermore, in this paper a more challenging and practical ZSL problem is considered where the original semantics are from natural language with irrelevant words instead of clean semantics that are widely used in previous work. Specifically, a multi-modal consistent bidirectional generative adversarial network is trained to handle unseen instances by leveraging noise in the natural language. A forward one-to-many mapping from one text description to multiple visual features is coupled with an inverse many-to-one mapping from the visual space to the semantic space. Thus, a multi-modal cycle-consistency loss between the synthesized semantic representations and the ground truth can be learned and leveraged to enforce the generated semantic features to approximate to the real distribution in semantic space. 
    %Notably, in the inverse mapping, apart from the adversarial loss, a classifier is trained on the semantic representation to guarantee that class-specific information is fully uncovered from the natural description. 
    Extensive experiments are conducted to demonstrate that our method consistently outperforms state-of-the-art approaches on natural language-based zero-shot learning tasks.
    
\end{abstract}

%%%%%%%%% BODY TEXT
\section{Introduction}

Over the past few years, deep learning techniques have remarkably boosted the performance of object classification tasks. This success is attributed to the availability of enormous amount of data for training. However, it is unlikely to collect training data for every class in the real world. In order to tackle such a problematic situation, zero-shot learning \cite{Label-embedding} as a promising solution to recognize new categories with limited training categories has been widely researched recently. The early ZSL aims to find an intermediate semantic representation to transfer the knowledge learned from seen categories to unseen ones. Recently, generative methods are further studied to directly synthesize unseen visual features.

Existing generative approaches synthesizing visual features for unseen classes with Generative Adversarial Networks (GANs) \cite{zhu2018generative,chen2018zero, xian2018feature} are proposed to address the data missing problem of unseen classes. Generally, the class semantic prototypes together with some noises are fed into these generative models to enforce the synthesized visual features as realistic as the real visual features. Once the models are trained, plausible visual features can be generated given semantic prototypes of unseen classes. However, merely based on adversarial losses, the visual feature generators cannot guarantee that the synthesized features truthfully reflect the corresponding semantics.

On the other hand, existing ZSL methods generally assume that the semantic representations are available and clean. However, it requires domain experts to manually annotate attributes. Furthermore, collecting hundreds of attributes as semantic representations for each category is extremely time-consuming and tedious. In contrast, online noisy text descriptions, e.g., Wikipedia articles, are much easier to collect. And, more excitingly, they are free.

%---------------------------------------------------
 \begin{figure}[t]
\centering
\includegraphics[width=85mm]{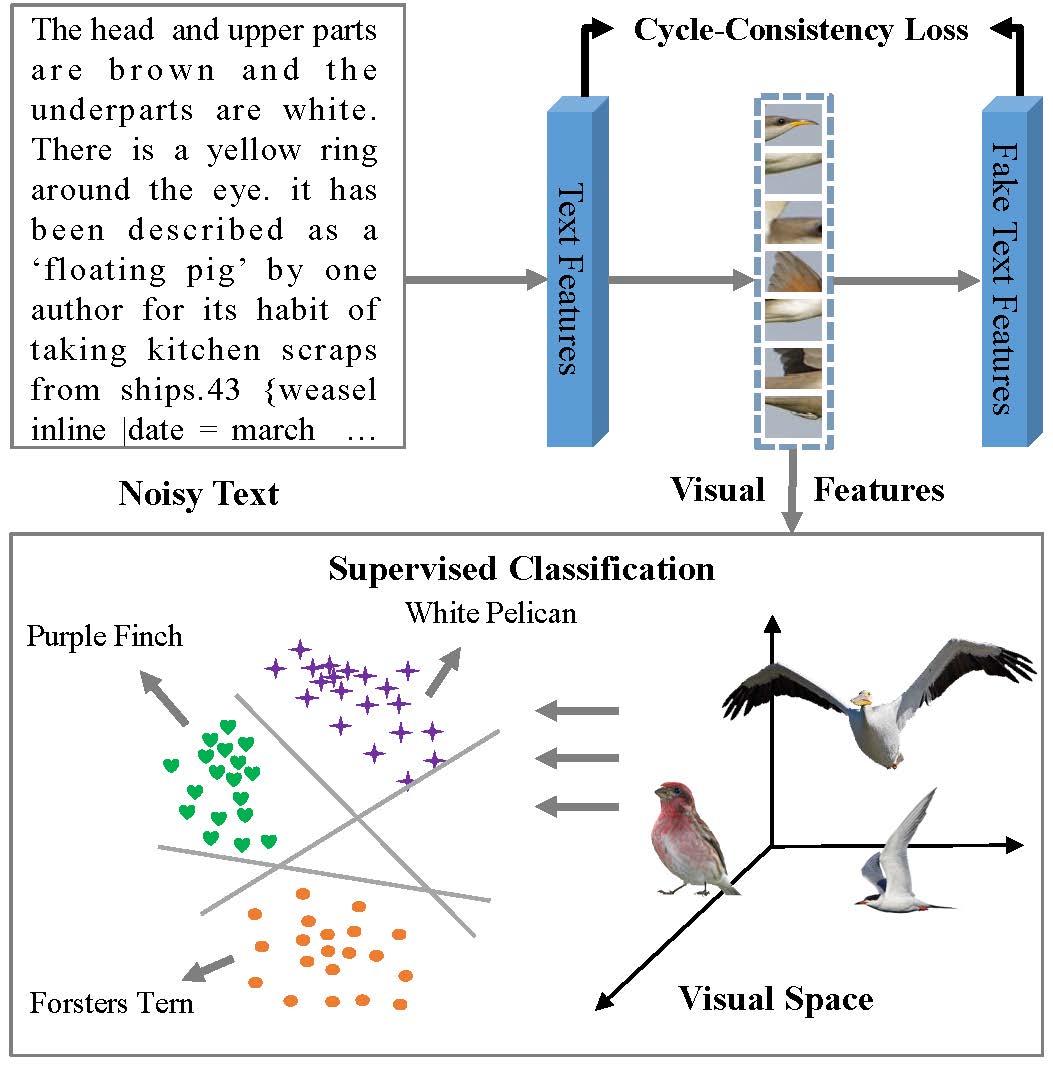}
\caption{%An illustration of our proposed approach. 
Our CANZSL model leverages cycle-consistent adversarial networks to synthesize realistic visual features from natural language. The problem is simplified as training a supervised classifier to predict the image labels.}
\end{figure}
%---------------------------------------------------

To address these two issues, we propose a novel cycle model, as shown in Fig. 1, to synthesize realistic and discriminative visual features from noisy text representations. A supervised classifier can be simply trained to predict the labels of unseen objects with the synthesized visual features, which are sampled from unseen semantic representations. Specifically, there are three main components in our architecture as shown in Fig. 2. Firstly, a fully connected layer is used to denoise and embed the natural language into pure textural representations. Secondly, a WGAN \cite{arjovsky2017wasserstein} is utilized to leverage the Wasserstein distance between the real and generated visual feature distributions, considering its demonstrated capability of extinguishing mode collapse. Acting as the main component in our framework for visual feature generating, the WGAN is able to generate diverse visual features. In addition to the weight clipping on the discriminator for satisfying the Lipschitz constraint, we further train a classification network on the discriminator. The classifier is adopted to categorize both the synthesized and the real visual features into correct classes. The classification loss also regularizes the generated visual features to be as much discriminative as the real visual features. Thirdly, in order to guarantee that the synthesized visual features can accurately reflect the corresponding semantics, we adopt an inverse adversarial network to convert the synthesized visual features back to textual features. By applying the cycle-consistent loss between the output text features from the inverse GAN and the input text features to the forward GAN, the inverse GAN collaboratively boosts the forward GAN to capture the underlying data structure. Besides, in the inverse discriminator we train a classifier with the same manner of the visual feature classification. We argue that the textual feature classification loss prevents the embedding FC layer from losing semantic information while suppressing the noise.

The main contributions of this paper can be summarized as follows:

1) We propose a novel structure called cycle-consistent adversarial networks for zero-shot learning, which is capable of synthesizing missing visual features for unseen classes from noisy Wikipedia articles.

2) Different from existing approaches, which only deploy a single GAN to learn semantic to visual mapping, our model consists of two symmetric GANs, a forward GAN and an inverse GAN. They collaboratively promote each other by the constraint of the cycle-consistency loss, the adversarial loss, and the classification loss.

3) The performance of the proposed CANZSL is verified on two tasks: zero-shot recognition and generalized zero-shot learning. Extensive experiments on two benchmark datasets, CUB and NAB, demonstrate that the proposed method consistently outperforms state-of-the-art methods.

%The remainder of this paper is organized as follows. Sec. 2 gives a brief discussion of related work and highlights the difference of our model. Sec. 3 describes the problem formulation, model architecture and training procedure. Experiments are presented in Sec. 4, and Sec. 5 concludes this paper.

%-------------------------------------------

\section{Related Work}

\subsection{Zero-shot Learning}

Zero-shot learning aims to overcome the issue of increasing difficulty in collecting data for a large amount of categories. Most exciting methods for zero-shot learning are attribute-based visual recognition \cite{ lampert2009learning, romera2015embarrassingly,norouzi2013zero,jiang2017learning} where the object attributes work as an intermediate feature space that transfer knowledge across object categories.

However, unlike well-specified attribute representations, the real world data is mostly natural language, \eg, Wikipedia articles. In this case, there is another research direction that explores zero-shot learning using online text articles. 
Elhoseiny \etal\cite{elhoseiny2013write} proposed an approach based on regression and domain adaptation that utilizes unpaired textual descriptions and images. 
%Elhoseiny \etal\cite{elhoseiny2013write} proposed an approach that combines domain transfer and regression to predict visual classifiers from a TF-IDF textual representation.
Bo \etal\cite{lei2015predicting} took advantage of deep convolutional neural network architecture and utilized latent features from different layers, resulting in a remarkable improvement on zero-shot recognition. 
%Bo \etal\cite{lei2015predicting} adopted deep neural networks to predict convolutional classifiers, leading to a noticeable improvement on zero-shot classification.
Qiao \etal\cite{qiao2016less} proposed a noise suppression technique for noisy signal in text based on $l_{2,1}-norm$ and learn a function to match the text document and the visual features. Further, they also analyzed in-depth that which particular information in documents is useful for zero-shot learning.
%Qiao \etal\cite{qiao2016less} revisited the importance of regularization on zero-shot learning. Interestingly, they also showed that attribute-based formulation \cite{romera2015embarrassingly} is able to achieve competitive performance when applied to text description. 
%It is further demonstrated that the noise in the text descriptions could be suppressed by encouraging group sparsity on the connections to the textual representations. 
%In our work, we directly suppress the textual noise by a fully connected layer as shown in Sec. 3.2.2. This simple implementation yields a competitive performance.

Another strategy for zero-shot learning converts the zero-shot problem to a traditional supervised classification task by sampling realistic visual features for unseen categories \cite{guo2017synthesizing, guo2017zero, zhu2018generative}.
%long2017zero,
Guo \etal\cite{guo2017synthesizing} estimated the probability distribution of unseen classes from the knowledge acquired from seen classes, and trained supervised classifiers according to the samples that are synthesized based on the distribution. They \cite{guo2017zero} also proposed an approach to synthesize images directly from seen class probability distribution where the noise from images undoubtedly cause side effects. GAZSL \cite{zhu2018generative} adopted a single GAN model to synthesize visual features from semantics and achieved state-of-the-art performance. 

%Guo \etal\cite{guo2017synthesizing} used Gaussian distribution priors for visual features in each class and estimated the distribution of unseen class as a linear combination of those seen classes.
%Long \etal \cite{long2017zero} adopted one-to-one mapping strategy and synthesized visual features by mapping attributes of classes to the visual space.
 %In comparison, our adversarial based approach, do not require any prior assumption of data probability distribution and are able to generate unlimited amount of synthesized samples. 

 \begin{figure*}[t]
\centering
\includegraphics[width=170mm]{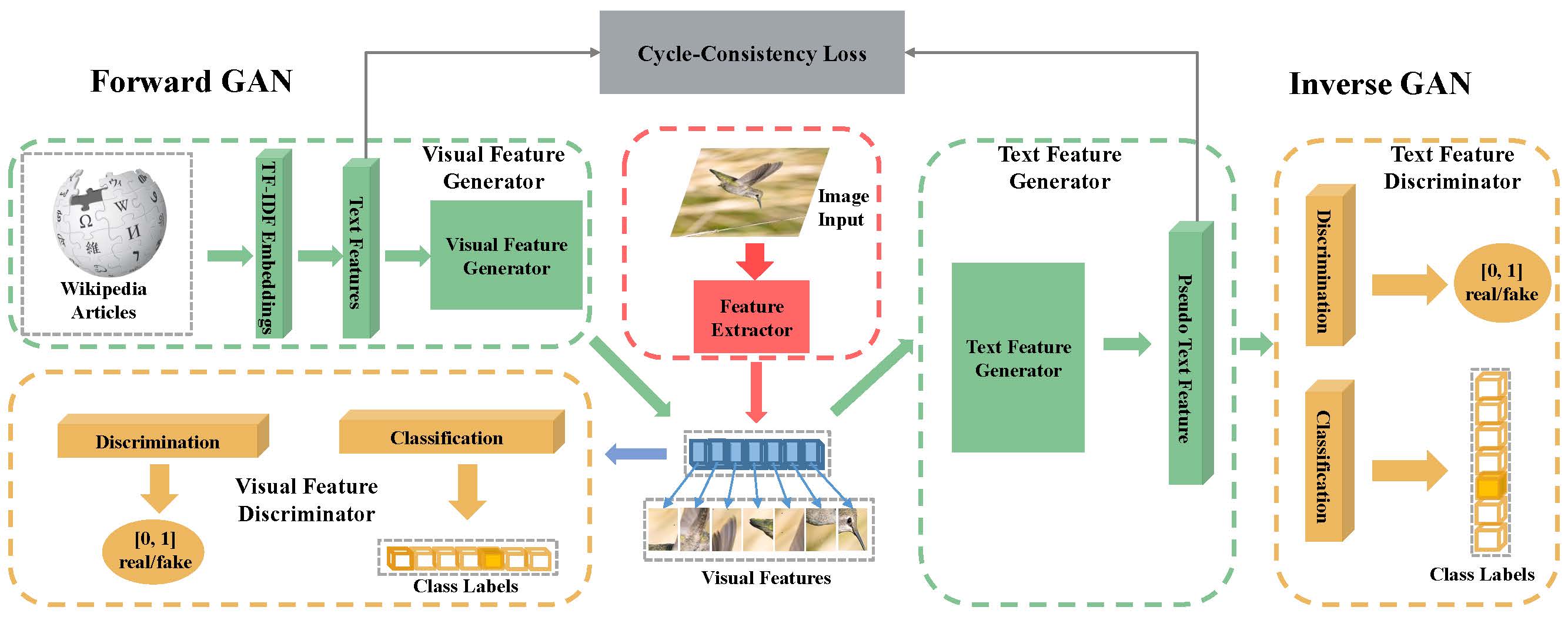}
\caption{Model overview. Given input images, our approach first extracts deep visual features as real samples throughout the training procedure (the pink part). The forward GAN (on the left hand side) synthesizes realistic and discriminative visual features from the TF-IDF embeddings of Wikipedia articles, whereas the Inverse GAN takes as input the generated visual features to reconstruct into text features again. Then the cycle-consistent loss can be applied to regularize the forward GAN to uncover semantics from the noisy TF-IDF features.}
\vspace{-5pt}
\end{figure*}
\subsection{Generative Adversarial Networks}
\vspace{-5pt}
Generative Adversarial Networks (GANs) \cite{goodfellow2014generative,zhao2016energy} have demonstrated favorable performance on image generation \cite{xu2018attngan,denton2015deep}, image editing \cite{zhu2016generative}, and representation learning \cite{radford2015unsupervised,salimans2016improved}. A GAN consists of a generator and a discriminator, and the idea behind is training the generator that can fool the discriminator to confuse the distributions of the generated and true samples. Theoretically, the training procedure can allow the generator to perfectly model the data distribution. However, it is usually hard to train a GAN, and mode collapse is known as a common issue of GANs due to the lack of explicit constraint in the learning objective. There are many methods \cite{zhao2016energy, mao2016multi, gulrajani2017improved,arjovsky2017wasserstein} recently proposed to stabilize the training procedure of GANs and mitigating model collapse issues, via using alternative objective functions. WGAN \cite{arjovsky2017wasserstein} introduced the Wasserstein distance among distributions as the objective function, and showed its capability of mitigating mode collapse. They applied weight clipping on the discriminator to allow the Lipschitz constraint. The improved WGAN \cite{gulrajani2017improved} used an additional gradient penalty instead of weight clipping to get rid of side effects in WGAN. We adopt WGAN with conditional information as the generative model in our proposed architecture and further use the gradient penalty technique to accelerate the convergence of WGAN. 

In a conditional setting, compared with a standard GAN, conditional GANs (cGANs) \cite{mirza2014conditional,chen2017show} take additional information vectors (e.g., textual descriptions) as input to both generator and discriminator. The additional information enables the generator to synthesize samples corresponding to the given condition. Auxiliary Classifier GAN (ACGAN) \cite{odena2017conditional} adopted extra classification information in the discriminator, which encourages the generator to synthesize samples based on the class labels as well. In the proposed model, the forward GAN and the inverse GAN take text features and visual representations as input, respectively. Also, we train two classifiers in each discriminator so that the classification information can be preserved.

\subsection{Cycle Architecture}
A cycle consistency error is proposed in addition to adversarial losses by Zhu \etal ~\cite{CycleGAN} to tackle the problem of lacking image-to-image translation training data. It is worth mentioning that the CycleGAN can be viewed as training two autoencoders \cite{hinton2006reducing}; each has a opposite internal structure to the other. Such a setup can also be seen as a special case of "adversarial autoencoders" \cite{makhzani2015adversarial}, which trains the bottleneck layer of an autoencoder by using an adversarial loss to approximate an arbitrary target distribution. 

The CycleGAN \cite{CycleGAN} framework is then widely adopted by a number of works. Cycada \cite{hoffman2017cycada} utilized the cycle model to perform various applications, e.g., digit adaption, cross-sense adaption. CamStyle \cite{zhong2018camera} is a camera style adaption approach proposed to conduct person re-identification tasks based on CycleGAN. It is also applied to cross-model retrieval \cite{wu2019cycle}, where a number of hash functions are learned to enable translation between modalities while using the cycle-consistency loss to enforce the correlation between outputs and original inputs. The above mentioned methods demonstrated the superiority of the cycle architecture in various tasks. Inspired by this observation, we apply the cycle architecture in zero-shot learning from natural language. 

%-------------------------------------------

\section{Approach}
We first introduce the problem formulation and then discuss in detail the proposed cycle-consistent adversarial networks for ZSL. Lastly, we illustrate our training procedure and how we conduct our zero-shot recognition task.

\subsection{Problem Formulation}
Given a batch of seen instances defined by $N^s$ triplets\space $\{(x_{i},\alpha_{i},y_{i})\}_{i=1}^{N^s}$, where $x_{i} \in \chi^s$ denotes the image features, $\alpha_{i}^s$ and $y_{i}^s$ represent the corresponding TF-IDF vector from Wikipedia articles and the associated one-hot class label respectively. Note that the seen instances $S$ and the unseen instances $U$ are disjointed $S \cap U = \varnothing$. In the test phase, it is assumed that the visual feature $x_{i}^u$ and TF-IDF vector $\alpha_{i}^u$ of a new category are provided, ZSL aims to predict the category label $y_{i}^u$. 

\subsection{Model Architecture}
Our model mainly comprises two components: 
%a visual feature extractor $E$, 
a forward visual feature synthesis network $F = \{G_{1}, D_{1}\}$ and an inverse text feature generation network $V = \{G_{2}, D_{2}\}$.

~\
\vspace{-20pt}
\subsubsection{Visual Feature Synthesis Network} ~\
Our Forward Network $F = \{G_{1}, D_{1}\}$ for visual feature synthesis is a conditional WGAN with auxiliary information. Specifically, it comprises a generator $G_{1}$ and a discriminator $D_{1}$. We simply use a fully connected layer for text embedding in the generator and regularize the visual features with the mean value in each classes, so that the distances between categories can be preserved in visual space.

\textbf{Visual Feature Generator $G_{1}$}:  Given the Term Frequency-Inverse Document Frequency (TF-IDF) features $\alpha$ from the seen natural language descriptions, we first use a fully connected layer as the text encoder $\psi$to generate text embedding $a \leftarrow$ $\psi(\alpha)$. The knowledge-distilled text embedding is then concatenated with a random noise distribution $z \in \mathbb{R}^z$ sampled from Gaussian distribution $N(0,1)$. The next training step is feeding the concatenated vector $[a,z]$ into two fully connected layers, and each followed by activation functions - Leaky ReLU and Tanh respectively. So far, the fake visual features and the text features $[\hat{x}$, $s]$ = $G_{1}(\alpha, z)$ are generated and the objective of the feature generation network can be formulated as:
\begin{equation}
\begin{gathered}
  \mathcal{L}_{G_{1}} = - \mathbb{E}_{z \sim p_{z}}[D_{1}(G_{1}(\alpha,z,\theta),w)]   \\ + \mathcal{L}_{cls_{1}}(G_{1}(\alpha,z,\theta)),
\end{gathered}
\end{equation}

\noindent where the $D_{1}$ represents Wasserstein loss \cite{arjovsky2017wasserstein} and the $\mathcal{L}_{cls_{1}}$ is the visual feature classification loss according to category labels, which will be introduced in detail in the Discriminator. $\theta$ and $w$ signify the parameters in the generator and the discriminator.

\textbf{Visual Feature Discriminator $D_{1}$}: The synthesized visual features from visual feature generator $G_{1}$ and the real image features 
%extracted from E 
are fed into $D_{1}$. After the input visual features passing through a fully connected layer and activation function ReLU, we use a fully connected layer to distinguish if the input features are real or not, and simultaneously use another fully connected layer to classify the input image features into correct categories. Introducing classification loss in discriminator has shown its promising effects in Auxiliary Classifier GAN \cite{odena2017conditional}. The objective function of the visual feature discrimination network can be defined as:

\begin{equation}
\begin{gathered}
  \mathcal{L}_{D_{1}} =\frac{1}{2}(\mathcal{L}_{cls_{1}}(G_{1}(\alpha,z,\theta))
  + \mathcal{L}_{cls_{1}}(x)) + \mathcal{L}_{GP_{1}} \\ + \mathbb{E}_{z \sim p_{z}}[D_{1}(G_{1}(\alpha,z,\theta),w)] - \mathbb{E}_{x \sim p_{data}}[D_{1}(x,w)]     
  ,
\end{gathered}
\end{equation}

\noindent where the first two terms are visual feature classification losses, we experimentally set the coefficient for fake features as $\frac{1}{2}$ since it works stably over different evaluations. $\mathcal{L}_{GP_{1}}$ is the gradient penalty term for applying the Lipschitz Constraint: $\mathcal{L}_{GP_{1}} = \lambda (||\triangledown_{\tilde{x}} D_{1}||_{2} - 1)^2$ where the $\tilde{x}$ is the linear interpolation of the fake feature $\hat{x}$ and the real feature $x$. The last two $D_{1}$ loss functions calculate Wasserstein distance of the fake and the real visual features,

\subsubsection{Text Feature Generation Network} ~\
Similar to the visual feature generation network, our proposed inverse Text Feature Generation Network $I = \{G_{2}, D_{2}\}$
consists of a text feature generator $G_{2}$ and a text feature discriminator $D_{2}$. The main contribution of this model
to the overall architecture is to provide reconstructed text features for calculating cycle-consistency loss, and further
regularize the text embedding layer to generate accurate text features by introducing text feature classification loss.

\textbf{Text Feature Generator $G_{2}$}: Given the synthesized image features $\hat{x}$ from $G_{1}$, the goal of $G_{2}$ is to generate realistic text features. The input visual features are around 3500 dimensions, concatenated with a 100 dimension random noise $z \in \mathbb{R}^z$ sampled from Gaussian distribution $N(0,1)$. The concatenated vectors are then fed into two fully connected layers together with Leaky-ReLU and Tanh activation functions respectively. The synthesized text features $\hat{\alpha}$ = $G_{2}(\hat{x}, z,\delta)$ are so far prepared for applying cycle-consistent constraint. The objective function of text feature generator $G_{2}$ can be formulated as:
\begin{equation}
\begin{gathered}
  \mathcal{L}_{G_{2}} = - \mathbb{E}_{z \sim p_{z}}[D_{2}(G_{2}(\hat{x},z,\delta),\zeta)]  ~\\ 
  + \mathcal{L}_{cls_{2}}(G_{2}(\hat{x},z,\delta),\zeta),
\end{gathered}
\end{equation}
where the $D_{2}$ and the $\mathcal{L}_{cls_{2}}$ represent Wasserstein loss \cite{arjovsky2017wasserstein} and the text feature classification loss corresponding to category labels, respectively. $\delta$ and $\zeta$ represents the weights in $G_{2}$ and $D_{2}$.

\textbf{Text Feature Discriminator $D_{2}$}: Once the reconstructed text features are mapped back from synthesized visual features, they are processed through a fully connected layer with ReLU activator. Afterwards, same as $D_{1}$ we simply use a fully connected layer for distinguishing the text feature fidelity and another fully connected layer to classify the text features into different categories. The objective function of $D_{2}$ is defined as:
\begin{equation}
\begin{gathered}
  \mathcal{L}_{D_{2}} = \frac{1}{2}(\mathcal{L}_{cls_{2}}(G_{2}(\hat{x},z,\delta))  + \mathcal{L}_{cls_{2}}(s)) + \mathcal{L}_{GP_{2}}\\ + \mathbb{E}_{z \sim p_{z}}[D_{2}(G_{2}(\hat{x},z,\delta),\zeta)] -
  \mathbb{E}_{x \sim p_{data}} [D_{2}(x,\zeta)],
\end{gathered}
\end{equation}
\noindent where the first two terms are text feature classification losses, which enforce the text feature embedding to be as well discriminative as the visual features. $\mathcal{L}_{GP_{2}}$ is the gradient penalty computed with the same manner of the visual feature discriminator and the last two terms are Wasserstein distance of the fake and the real visual features,

~\
\vspace{-10pt}
\subsubsection{Cycle-Consistency Loss}~\
In theory, learning a forward mapping and a inverse mapping by adversarial losses is able to produce outputs identically distributed as real features \cite{goodfellow2014generative}. However, even if the forward mapping is conditioned on the seen semantic features, there is no guarantee that the synthesized visual features capture textual features. In order to address this issue, we introduce cycle-consistency loss $\mathcal{L}_{cyc}$ to regularize the visual feature generator $G_{1}$ being able to synthesize visual features with semantic information preserved. Once the reconstructed text features are generated by $G_{2}$, the cycle consistency loss is computed to update weights on both $G_{1}$ and $G_{2}$. We also argue that the cycle-consistency loss in our architecture promote the text encoder $\psi$ as well, which is included in $G_{1}$. It is defined as:

\begin{equation}
\begin{gathered}
  \mathcal{L}_{cyc} = \lambda \frac{1}{N^b}\sum^{N^b}_{n=1}||G_{2}(G_{1}(\alpha,z,\theta),z,\delta) - s||^2,
\end{gathered}
\end{equation}

\noindent where $\lambda$ is the coefficient, the $N^b$ denotes the batch size, and $s$ is the text feature from text encoder $\psi(\alpha)$. The cycle-consistency loss is essentially the mean squared error between the reconstructed textual features from $G_{2}$ and the real ones directly extracted from noisy text descriptions $\alpha$.

\subsection{Training Procedure} 
%To train the overall model, we consider visual-semantic feature pairs as joint observation. Visual features are either generated by our visual feature generator $G_{1}$ or extracted from visual feature extractor $E$, whereas text features are either generated by our text feature generator $G_{2}$ or encoded by the fully connected layer in $G_{1}$. We train two discriminators $D_{1}$ and $D_{2}$ separately to distinguish the reality and classify the object category of the synthesized visual features and text features respectively. We show our training procedure in Algorithms 1. In each iteration, these two discriminators are updated $n_{d}$ times (lines 2 - 7 and line 17 -22), and the both generators are optimized for 1 step ( line 8 - 16 and line 23 - 16). In addition, the visual pivot loss (line 14) and cycle-consistency loss (line 27 - 29) are applied once in each iteration respectively.

To train the overall model, we consider visual-semantic feature pairs as joint observation. Visual features are either generated by our visual feature generator $G_{1}$ or from the ground truth provided in the datasets, whereas text features are either generated by our text feature generator $G_{2}$ or encoded by the fully connected layer in $G_{1}$. The visual features and text features are further introduced in section 4.1. We train two discriminators $D_{1}$ and $D_{2}$ separately to distinguish the reality and classify the object category of the synthesized visual features and text features respectively. 
%We show our training procedure in Algorithms 1. 
In each training iteration, two discriminators are updated 5 times, and the both generators are optimized for 1 step. In addition, we follow a training technique from \cite{zhu2018generative} that regularize the generated visual features to be consistent with the cluster centre of the corresponding object class. Lastly, the cycle-consistency loss are applied once in each iteration by calculating the gap between the generated text features and the text features that directly extracted from natural language.

\subsection{Zero-Shot Recognition} 
Given unseen semantic descriptions and random noise $z$ from Gaussian distribution, our trained model can synthesize infinite number of visual features with randomly sampled $z$. The process can be formulated as following:
\begin{equation}
\begin{gathered}
  x_{u} = G_{1}(\alpha_{u},z,\theta)
\end{gathered}
\end{equation}
Once the visual features are synthesized, the zero-shot learning problem becomes a traditional supervised classification problem. For simplicity, we adopt k-nearest neighbor algorithm (K-NN) to conduct this supervised task.

%-------------------------------------------

\section{Experiments}
\subsection{Experiment Setting}

\textbf{Datasets}: We conduct experiments on two bird datasets: CUB-200-2011 (CUB) \cite{CUB} and North America Birds (NAB) \cite{NAB}. The CUB dataset consists of 11,788 images from 200 bird species, and NAB is a significantly larger dataset of birds with 1011 categories and 48,562 images. NAB dataset forms a hierarchy of bird classes, including 555 leaf nodes and 456 parent nodes. The images in NAB are associated with leaf nodes. Elhoseiny \etal ~\cite{Elhoseiny_2017} captioned both datasets with the Wikipedia article. Due to the lack in the Wikipedia articles for some subtle division of classes, some subclasses are merged and 404 classes of birds are yielded with corresponding Wikipedia articles.

%Besides, Elhoseiny \etal ~\cite{Elhoseiny_2017} designed two kinds of splitting schemes, based on how close the seen classes are to the unseen classes: 
Besides, there are two splitting designs according to the relationship between seen categories and unseen categories:
Super-Category-Shared splitting (SCS) and Super-Category-Exclusive splitting(SCE). 
%For SCS, unseen classes are deliberately picked on the condition that there exists seen classes with the same Super-Category. In this scheme, the relevance between seen and unseen classes is very high. 
In SCS, unseen categories are chosen to share same super-class with the seen categories. In result, the relevance in this design between seen categories and unseen categories is relatively high. 
%On the contrary, in SCE, all classes under the same category as unseen classes would either belong to the seen or the unseen classes. 
In contrast, all categories in SCE belong to the same super-class are split into either seen or unseen categories.
Intuitively, zero-shot recognition performance should be better in SCS-split than SCE-split. Conventional ZSL methods\cite{akata2016multi, akata2015evaluation, qiao2016less, romera2015embarrassingly} use SCS-split only, whereas we use both splits to validate our approach.

%In order to drastically demonstrate the superiority of our model, we follow the experiment setting with other competing methods.
\textbf{Text Feature}:
Elhoseiny \etal ~\cite{Elhoseiny_2017} collected raw Wikipedia articles for CUB and NAB datasets. They first tokenized the Wikipedia articles into words and got rid of the full stops. In order to weight the terms in the dataset appropriately, Term Frequency-Inverse Document Frequency (TF-IDF) \cite{salton1988term} is adopted to extract text feature vectors. The dimensionality of TF-IDF feature for dataset CUB \cite{CUB} and NAB \cite{NAB} are 11,083 and 13,585 respectively. However, the TF-IDF is further embedded into a lower dimension textual representations for suppressing the noise.

\textbf{Visual Feature}:
Elhoseiny \etal ~\cite{Elhoseiny_2017} prepared visual features from images in the two bird datasets with Visual Parts CNN Detector/Encoder (VPDE). Input images are reshaped to 224$\times$224 and detected by fast-RCNN framework with VGG16. The detected parts are then fed to the VPDE network, where 512-dimensional feature vectors are extracted for each semantic part. For dataset CUB, seven semantic parts are used to train the VPDE network. Due to the lack of annotations for the "leg" part in the NAB dataset, we use only six visual parts without the "leg" part. The dimensionality of visual features for CUB and NAB are 3582 and 3072 respectively.

\textbf{Implementation Details}:
Theoretically, the synthesized text features from the inverse GAN provides multi-modal constraint information to optimize the forward visual feature generation model. Thus, if the reconstructed text features from inverse GAN are accurate throughout the training process of the forward GAN, the cycle-consistent training should not only converge stably but also faster.

However, there are three reasons why we choose not to pretrain our inverse model. First, note that our proposed cycle architecture has a low complexity with only several fully connected layers. Even if we pretrain the inverse model, the convergence comes nearly same as training the whole model from scratch. Second, with the classification information introduced in both discriminators, the model usually finds the optimal gradient extremely quickly. Last but not least, with a large $\lambda$ for cycle-consistency loss (we set as 10 in our experiments), the forward and the inverse networks both promote each other collaboratively.

In order to compare our approach with GAZSL \cite{zhu2018generative} to show the superiority of our cycle-consistent architecture, we follow the same setting in the forward visual feature generation GAN as GAZSL. The batch size is fixed as 1000, and the learning rate, as shown in Algorithm 1, is set as 0.0001. We use Adam optimizer \cite{kingma2014adam} with $\beta_{1}$ as 0.5 and $\beta_{2}$ as 0.9 respectively. All experiments are conducted on a server with 16 Intel(R) Xeon(R) Gold 5122 CPUs and 2 GeForce RTX 2080 Ti GPUs.

\begin {table}[t]
\caption {Top-1 accuracy (\%) on CUB and NAB datasets with two split settings. }

\begin{center}

\begin{tabular}[t]{ l | c | c | c | c }
%\specialrule{.1em}{.00em}{.00em}
\hline
   & \multicolumn{2}{|c|}{CUB}  & \multicolumn{2}{|c}{NAB} \\
  \hline
    Methods              & SCS   & SCE   & SCS   & SCE     \\
%\specialrule{.1em}{.00em}{.00em}
  MCZSL\cite{akata2016multi}                 & 34.7  & -     & -     & -       \\
  WAC-Linear\cite{elhoseiny2013write}            & 27.0  & 5.0   & -     & -       \\
  WAC-Kernel  \cite{elhoseiny2017write}          & 33.5  & 7.7   & 11.4  & 6.0     \\
  ESZSL    \cite{romera2015embarrassingly}             & 28.5  & 7.4   & 24.3  & 6.8     \\
  SJE  \cite{akata2015evaluation}                 & 29.9  & -     & -     & -       \\
  ZSLNS \cite{qiao2016less}                 & 29.1  & 7.3   & 24.5  & 6.8     \\
  SynC$_{\textit{fast}}$ \cite{changpinyo2016synthesized}  & 28.0  & 8.6   & 18.4  & 3.8     \\
   SynC$_{\textit{OVO}}$ \cite{changpinyo2016synthesized}  & 12.5  & 5.9   & -     & -       \\
  ZSLPP\cite{Elhoseiny_2017}                 & 37.2  & 9.7   & 30.3  & 8.1     \\
  GAZSL\cite{zhu2018generative}                 & 43.7  & 10.3  & 35.6  & 8.6     \\
  \hline
  CANZSL                & \textbf{45.8}  &   \textbf{14.3}   & \textbf{38.1}     &   \textbf{8.9}    \\
%\specialrule{.1em}{.00em}{.00em}
\hline
\end{tabular}

\end{center}
\vspace{-12pt}
\end {table}

\subsection{Performance evaluation}
%\textbf{Baselines and Competing Methods:}
Experiments are conducted on both SCE and SCS splits of the two bird datasets CUB and NAB to evaluate our approach. We compare our approach with other eight state-of-the-art algorithms: 
GAZSL \cite{zhu2018generative}, 
ZSLPP\cite{Elhoseiny_2017}, 
SynC\cite{changpinyo2016synthesized}, 
ZSLNS\cite{qiao2016less}, 
SJE\cite{akata2015evaluation},
ESZSL\cite{romera2015embarrassingly}, 
WAC\cite{elhoseiny2017write}, 
MCZSL\cite{akata2016multi}. 
The source code of GAZSL, ZSLPP, ESZSL, and ZSLNS are available online. For the rest of methods, we directly cite the highest scores reported in \cite{zhu2018generative}. For the attribute-based methods, we simply replace the attributes input with the textual features. ZSLPP and MCZSL extracts visual features from the semantic parts of birds. MCZSL simply adopted annotated semantic parts to supervise visual feature extraction during the testing stage. In comparison, our approach, ZSLPP and GAZSL used detected semantic parts in both training and testing phase. As a result, the performance of the final zero-shot classification is expected to degrade due to less accurate detection of semantic parts compared to manual annotation in MCZSL. Table 1 demonstrates the performance comparisons on CUB and NAB datasets. Generally, our method consistently outperforms the state-of-the-art methods. On the conventional split setting (SCS), our approach outperforms the runner-up (GAZSL) by a considerable gap: 2.1\% and 2.5\% on CUB dataset and NAB dataset, respectively. However, ZSL on SCE-split remains rather challenging. The fact that there is less relevant information between the training and testing set makes it extremely hard to transfer knowledge from seen classes to unseen classes. Although our method  improves the performance by 4\% on the CUB dataset, the improvement on NAB is merely 0.3\%. We will show a higher improvement on the general merit of ZSL in Sec 4.5 with two split settings. 

\begin {table}[t]
\caption {Effects of different components on zero-shot classification accuracy (\%) on CUB and NAB datasets with SCS split setting. }
\begin{center}
\scalebox{0.95}{
\begin{tabular}[t]{ l | c | c | c | c }
%\specialrule{.1em}{.00em}{.00em}
\hline
   & \multicolumn{2}{|c|}{CUB}  & \multicolumn{2}{|c}{NAB} \\
  \hline
  Methods             & \makecell{Text \\ feature}   & TF-IDF    & \makecell{Text \\ feature}  & TF-IDF      \\
%\specialrule{.1em}{.00em}{.00em}
\hline
CYC-only           & 45.1  & 44.8   & 36.5     & 35.9       \\
ADV-CYC-only         & 45.5  & 45.2   & 37.2  & 37.1    \\
CLA-CYC-only             & 45.3  & 44.9   & 37.1  & 36.7     \\

  \hline
  CANZSL                & \textbf{45.8}  &   \textbf{45.5}   & \textbf{38.1}     &   \textbf{37.3}    \\
%\specialrule{.1em}{.00em}{.00em}
\hline
\end{tabular}}

\end{center}
\end {table}

\subsection{Ablation Study}
We now report the ablation study of the effect of the cycle-consistency loss, the classification loss and the adversarial loss in the inverse GAN. We trained three variants of our model by only keeping the cycle-consistency loss, adversarial loss and classification loss, denoted as CYC-only, ADV-CYC-only, and CLA-CYC-only, respectively. In the case of CYC-only, the textual feature generator is merely updated by the cycle-consistency loss, whereas ADV-CYC is optimized by the cycle-consistency loss as well as an adversarial loss from the discriminator. Similarly, the CLA-CYC-only variant is optimized by the cycle-consistency loss and the classification loss. 

Table 2 shows the performance of each setting. It is clear that each component significantly contributes to the overall architecture. We also observe that with any proposed component, the performance of each variants is much higher than the runner-up method GAZSL shown in Table 1, which demonstrates the importance of each component. We argue that the adversarial loss and the classification loss are critically complementary to each other. The cycle-consistency loss can only ensure the mapping from synthesized textual features are accurately corresponding to the extracted ones from visual feature generation network. However, with the adversarial loss and classification loss applied on the pseudo textual feature generator, the text encoder in visual feature synthesis network is beneficial from the cycle-consistency loss by being forced to adapt to class label information.  

We investigate whether the cycle-consistency loss should be applied on the textual feature noisy or the TF-IDF representation of text description. As shown in Table 3, generally our method with cycle-consistency loss applied on textual features outperforms the one with cycle-consistency loss applied on noisy text TF-IDF representations. 

 \begin{figure}[t]
\centering
\includegraphics[width=85mm]{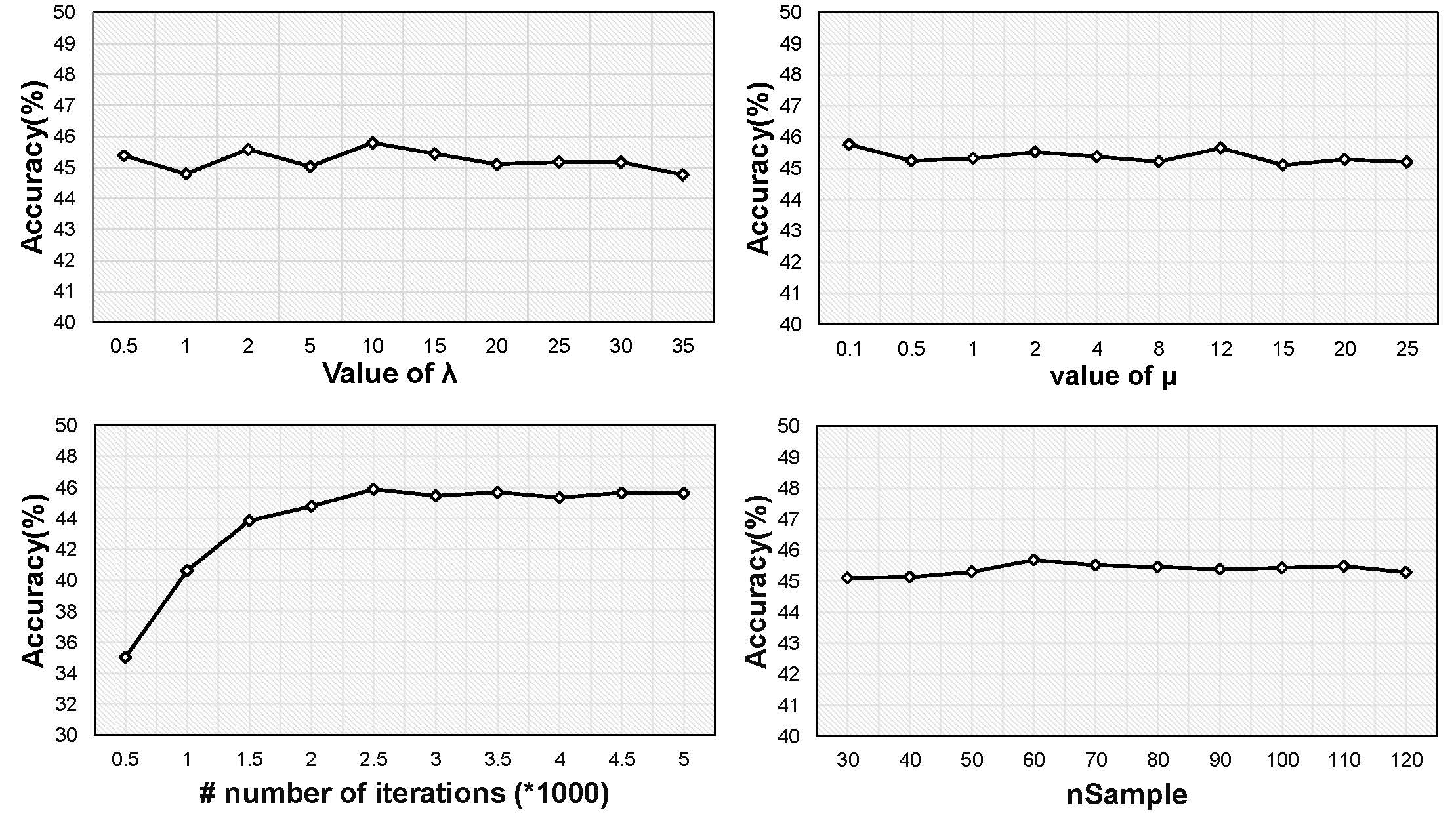}
\caption{Parameters sensitivity of the proposed method.}
\vspace{-10pt}
\end{figure}

\subsection{Parameters Sensitivity}
In order to investigate the most appropriate hyper-parameter values for the proposed CANZSL model, we compare and demonstrate the performance with various hyper-parameter values.
In our cycle architecture, we argue that the cycle-consistency loss is the most significant component according to the performance demonstrated in the ablation study. Even if the forward generator and the inverse generator are merely updated by cycle-consistency loss, we can outperform our baseline GAZSL \cite{zhu2018generative} by $1.4\%$. Here we demonstrate the performance comparison between various coefficients $\lambda$ for cycle-consistency loss. It is shown at the upper-left on Fig. 3 that when the $\lambda$ increases from 0.5 to 10, the performance is extremely unstable, and it decreases slowly when $\lambda$ is greater than 10. Intuitively, we argue that 10 is the best coefficient for cycle-consistency loss.

We also experimented on several different values $\mu$ for the coefficient of classification loss in the inverse network. Interestingly, there is no obvious trend as $\lambda$ in the upper-right graph on Fig. 3. As a result, our model is not sensitive to the hyper-parameter of $\mu$. Intuitively, we adopt the value 12 with the highest performance.

Even if our CANZSL model only involves a number of fully connected layers, which guarantee that the training usually converges drastically. From the lower-left line chart in Fig. 3, we can see the performance reaches 45.8$\%$ merely in 2500 iterations.. Afterwards, the performance keeps stable with the iteration goes up to 5000.

In testing phase, we uses different sampling numbers to evaluate our trained model. The result is shown in the lower-right in Fig. 3. We yield best performance when sampling 60 visual features.

\subsection{Results of the Generalized ZSL}
The conventional zero-shot learning categorizes text samples into unseen classes without seen classes in test phase, whereas the seen classes are usually more common than unseen classes. In this case, it is unrealistic to assume we will never encounter unseen objects during the test phase \cite{chao2016empirical}. Chao \etal ~\cite{chao2016empirical} recently proposed a more appropriate metric called Area Under Seen-Unseen accuracy Curve (AUSUC) that can evaluate generalized zero-shot learning (GZSL) approaches, by acknowledging that there is an inherent trade-off between recognizing seen classes and recognizing unseen classes.

In order to compare with the runner-up approach GAZSL, we directly cite the performance results reported in \cite{zhu2018generative}. We show the AUSUC results on both SCS split and SCE split in Table 3 and we observe that the proposed CANZSL approach performs particularly competitive under the more realistic generalized ZSL task. On dataset CUB and NAB and corresponding splits, our CANZSL obtains superior performance with a large margin against the competitors. The result indicates that our approach performs much better than other competitors on alleviating the issue of the seen-unseen bias under the generalized ZSL scenario. In other words, the proposed approach can improve the performances of unseen classes while maintaining the performances of seen classes. 
%In addition, we notice that the classification performance of unseen classes are much better than those of the overall of seen and unseen classes. In result, the synthesized pseudo visual features are unlikely to be as good as the real visual features. 

\begin {table}[t]
\caption {The performances (in \%) of the generalized ZSL on CUB and NAB datasets with two split settings. }
\begin{center}
\begin{tabular}[t]{ l | c | c | c | c }
%\specialrule{.1em}{.00em}{.00em}
\hline
   & \multicolumn{2}{|c|}{CUB}  & \multicolumn{2}{|c}{NAB} \\
  \hline
    Methods     & SCS   & SCE   & SCS   & SCE     \\
%\specialrule{.1em}{.00em}{.00em}
\hline
  WAC-Linear\cite{elhoseiny2013write}            & 23.9  & 4.9   & 23.5     & -       \\
  WAC-Kernel  \cite{elhoseiny2017write}          & 22.5  & 5.4   & 0.7  & 2.3     \\
  ESZSL    \cite{romera2015embarrassingly}             & 18.5  & 9.2   & 24.3  & 2.9     \\
  ZSLNS \cite{qiao2016less}                 & 14.7  & 4.4   & 9.2  & 2.3     \\
  SynC$_{\textit{fast}}$ \cite{changpinyo2016synthesized}  & 13.1  & 4.0   & 2.7  & 0.8     \\
  SynC$_{\textit{OVO}}$ \cite{changpinyo2016synthesized}  & 1.7  & 1.0   & 0.1     & -       \\
  ZSLPP\cite{Elhoseiny_2017}                 & 30.4  & 6.1   & 12.6  & 3.5     \\
  GAZSL\cite{zhu2018generative}                 & 35.4  & 8.7  & 20.4  & 5.8     \\
  \hline
  CANZSL                & \textbf{40.2}  &   \textbf{12.5}   & \textbf{25.6}     &   \textbf{6.8}    \\
%\specialrule{.1em}{.00em}{.00em}
\hline
\end{tabular}
\end{center}
\vspace{-20pt}
\end{table}

\subsection{ t-SNE Demonstration}
Fig. 4 demonstrates the t-SNE \cite{maaten2008visualizing} visualization of the real visual features and the synthesized visual features from unseen classes on CUB dataset. From the real samples in Fig. 4(a) we can see that some categories overlap with each other by a large degree, such as \textit{black billed cuckoo} and \textit{yellow billed cuckoo}. The overlapping also exists in the synthesized features as shown in Fig. 4(b). It is reasonable when considering these two bird classes only differentiate in colour. This insight observation indicates that the underlying data distribution is well captured in our model. Also, thanks to class label information involved, the synthesized features are extremely discriminative as they obviously distribute in separate clusters. 

\begin{figure}[t]
\subfigure[Real visual features]{
\includegraphics[width=0.4\linewidth, height=2.7cm]{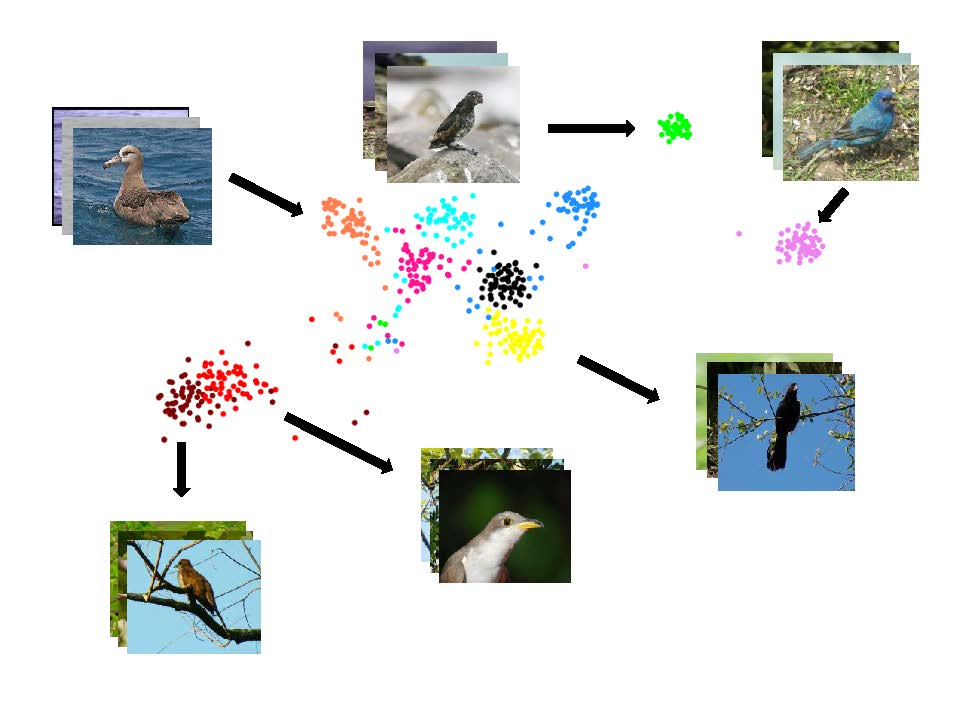} 
\label{fig:subim1}
}
\subfigure[Synthesized visual features]{
\includegraphics[width=0.55\linewidth, height=2.7cm]{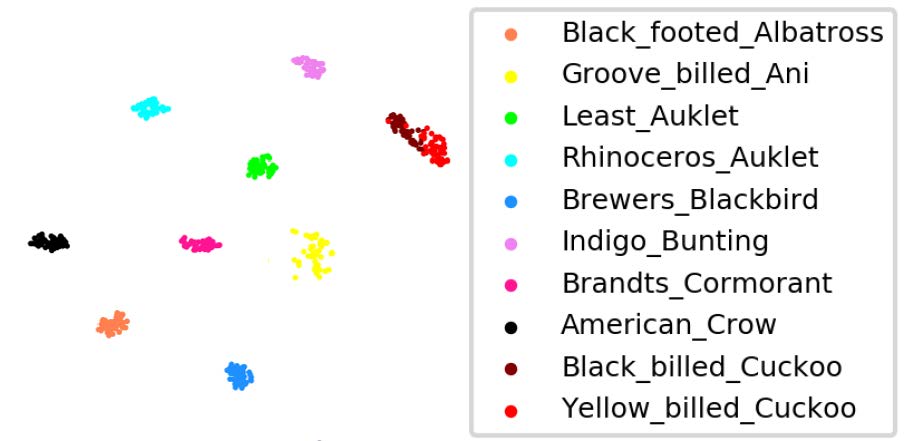}
\label{fig:subim2}
}
\vspace{5pt}
\caption{t-SNE visualization of features from random 10 unseen classes on CUB dataset. Each color represents a specific class label indicated on the right hand side of (b). (a) visualizes the visual representations directly extracted from the visual feature extractor E. (b) shows the visual features that inferenced from the trained forward GAN.}
\label{fig:image2}
\end{figure}

\subsection{Results of ZSL from Attributes}
As discussed in Sec. 2.1, it is more practical to conduct zero-shot learning from natural language. Further, we reported our superior performance on this task. In theory, during embedding the text description into textual features, it is unlikely to preserve non-trivial information thoroughly. Hence, in the same setting, our method should be able to perform better in zero-shot learning from attributes, which perfectly preserve all useful information. For zero-shot learning from attributes, the fully connected layer, which was adopted to suppress the text noise, is removed from the visual feature generator.

We demonstrate our performance in ZSL from attributes, and compare with four other state-of-the-art methods in Table 4. From the Table, we can notice that the proposed
method outperforms the state-of-the-art methods not only in ZSL from noisy text but from attributes as well. 
%Surprisingly, our method in zero-shot learning from attributes yields $\sim$ 10\% higher result than from noisy text.

\begin {table}[!h]
\caption {The performances (in \%) of ZSL from attribute-based semantic representation on CUB dataset. }
\begin{center}
\begin{tabular}[t]{ l | c | c }
%\specialrule{.1em}{.00em}{.00em}
   
  \hline
    Input              & Noisy Text   & Attributes       \\
%\specialrule{.1em}{.00em}{.00em}
    \hline
  ESZSL    \cite{romera2015embarrassingly}             & 28.5  & 53.9     \\
  SJE  \cite{akata2015evaluation}                 & 29.9  &53.9      \\

  SynC \cite{changpinyo2016synthesized}  & 28.0  & 55.6    \\
  GAZSL\cite{zhu2018generative}                 & 43.7  & 55.8     \\
  \hline
  CANZSL                & \textbf{45.8}  &   \textbf{56.5}       \\
%\specialrule{.1em}{.00em}{.00em}
\hline
\end{tabular}
\end{center}
\vspace{-20pt}
\end{table}
%-------------------------------------------

\section{Conclusions}
In this paper, we proposed novel cycle-consistent adversarial networks for ZSL from natural language, which leverage multi-modal cycle-consistency loss to regularize the visual feature generator to preserve semantics during training. An inverse GAN is added to reconstruct visual features back to textual representations. Experiments showed that our approach consistently performs favorably against the state-of-the-art methods not only on traditional ZSL, but on generative ZSL as well, with an outstanding capability of visual feature generation. We also showed in an ablation study that the adversarial loss, classification loss and cycle-consistency loss can promote the overall architecture collaboratively. Furthermore, we validated that our CANZSL is also able to perform well on the task of the ZSL from attributes. In our future work, we will also study how to optimize the testing phase and utilize the unseen class descriptions during training procedure.

%-------------------------------------------

%--------------------------------------------

{\small
\bibliographystyle{ieee}
\bibliography{egpaper_final.bbl}
}

\end{document}